\RequirePackage{fix-cm}
\documentclass[smallextended]{svjour3}       % onecolumn (second format)
\smartqed  % flush right qed marks, e.g. at end of proof
\usepackage{graphicx}
\usepackage[round]{natbib}
\usepackage{hyperref}
\renewcommand{\cite}{\citep}
\begin{document}

\title{Transformative Machine Learning%\thanks{Grants or other notes
}


\author{Ivan Olier         \and
        Oghenejokpeme I. Orhobor \and
        Joaquin Vanschoren \and
        Ross D. King
}


\institute{I. Olier \at
              Department of Applied Mathematics, Liverpool John Moores University \\
              Liverpool, UK, L3 3AF\\
              \email{I.A.OlierCaparroso@ljmu.ac.uk}           %  \\
           \and
           O. I. Orhobor \at
               School of Computer Science, University of Manchester \\
               Manchester, M13 9PL, UK \\
               \email{oghenejokpeme.orhobor@manchester.ac.uk}
            \and
            J. Vanschoren \at
                Eindhoven University of Technology \\
                5600MB Eindhoven, The Netherlands \\
                \email{j.vanschoren@tue.nl}
            \and
            R. D. King \at
                School of Computer Science, University of Manchester \\
                Manchester, M13 9PL, UK \\
                 Alan Turing Institute \\
                  London NW1 2DB, UK \\
                  National Institute of Advanced Industrial Science and Technology \\
                  Tokyo 135-0064, Japan \\
                  \email{ross.king@manchester.ac.uk}
}

\date{Received: date / Accepted: date}

\maketitle

\begin{abstract}
The key to success in machine learning (ML) is the use of effective data representations. Traditionally, data representations were hand-crafted. Recently it has been demonstrated that, given sufficient data, deep neural networks can learn effective implicit representations from simple input representations.  However, for most scientific problems, the use of deep learning is not appropriate as the amount of available data is limited, and/or the output models must be explainable. Nevertheless, many scientific problems do have significant amounts of data available on related tasks, which makes them amenable to multi-task learning, i.e. learning many related problems simultaneously. Here we propose a novel and general representation learning approach for multi-task learning that works successfully with small amounts of data. The fundamental new idea is to transform an input intrinsic data representation (i.e., handcrafted features), to an extrinsic representation based on what a pre-trained set of models predict about the examples.  This transformation has the dual advantages of producing significantly more accurate predictions, and providing explainable models. To demonstrate the utility of this transformative learning approach, we have applied it to three real-world scientific problems: drug-design (quantitative structure activity relationship learning), predicting human gene expression (across different tissue types and drug treatments), and meta-learning for machine learning (predicting which machine learning methods work best for a given problem). In all three problems, transformative machine learning significantly outperforms the best intrinsic representation.
\end{abstract}

\section{Introduction}
\label{sec:intro}
Machine learning (ML) is the branch of Artificial Intelligence (AI) that focuses on developing systems that can learn from experience. Rather than being explicitly told how to solve a problem, ML algorithms are able to learn from observations – induction \cite{russell2016artificial}. As ML algorithms have a generic ability to learn, rather than solve any particular problem, they are very widely applicable.
The application of ML to science has a long history. The pioneering work was the development of learning algorithms for the analysis of mass-spectrometric data \cite{buchanan1968heuristic}.  Now, the significance of ML to science has been generally recognized, and ML is being applied to a wide variety of different scientific areas, such as functional genomics \cite{king2009automation}, physics \cite{schmidt2009distilling}, drug discovery \cite{schneider2017automating}, organic synthesis planning  \cite{segler2018planning}, materials science \cite{butler2018machine}, and medicine \cite{esteva2017dermatologist}.
Probably the most exciting current area of machine learning is that of deep neural networks (DNNs) \cite{lecun2015deep, silver2016mastering, esteva2017dermatologist}. Thanks to advances in computer hardware and the availability of vast amounts of data, DNNs have been shown to be capable of such impressive tasks as beating World Champions at games such as Go \cite{silver2016mastering}, and diagnosing skin cancers better than human specialists \cite{esteva2017dermatologist}. In practice, however, DNNs are applicable only to a very small subset of scientific problems for which such large amounts of data are available. In addition, in most scientific problems, there is a requirement for human comprehensible models, while DNNs only provide black-box models. 

\subsection{Representation Learning}
\label{subsec:repr_learning}
The key to success in machine learning (ML) is the use of effective data representations. Almost all machine learning is based on representations that use tuples of attributes, i.e. the data can be put into a single table, with the examples as rows, and the attributes (descriptors) as columns.  An attribute is a proposition that is possibly true about an example. (Examples are described as tuples, and not vectors, as the order of the attributes does not matter - as long as it is the same for all the examples.)  
The attributes used to describe examples are intrinsic properties of the examples that are believed to be important: for example if one wished to learn about the effectiveness of a drug, then properties of its molecular structure may be useful attributes; similarly, if one wished to learn about chess positions, then the position of the white King might be a useful attribute. Typically, one attribute is singled out as the one we want to predict, and the other attributes contribute information to make this prediction.  If this attribute is categorical then the problem is a discrimination/classification task, if the attribute is a real number then the problem is a regression one. Here, we focus on regression problems.
The recent success of DNNs has been based on their ability to utilize multiple neural network layers, and large amounts of data, to learn how to convert raw input representations (e.g., image pixel values) into richer internal representations that are effective for learning.  This internal conversion has been especially successful in problems where the only available attributes are very simple and minimal, such as pixel colour, brightness, position, etc. Due to this ability to learn effective internal representations, DNNs have succeeded in domains that had previously proved recalcitrant to ML, such as face recognition and learning to play GO.  The archetypical case of this is face recognition, which was once considered to be intractable, but can now be solved with super-human ability on certain limited problems \cite{bengio2012deep}.

\subsection{Multi-task Learning and Transfer Learning}
\label{subsec:multitask}
The large amounts of data required for DNNs to learn a good representation is unfortunately not available for many scientific problems. Nevertheless, many scientific problems do often present themselves as sets of related problems, which taken together, provide significant amounts of data, e.g. learning quantitative structure activity relationships (QSARs) for related targets (proteins).
Multi-task learning \cite{caruana1997multitask} is the branch of machine learning in which related problems (called tasks) are learned simultaneously, with the aim to exploit similarities between the tasks and thus obtain improved performance \cite{ando2005framework, evgeniou2005learning}. The tasks are learned in parallel using a shared representation, so that what is learned from one task (e.g. one where more data is available) can also be used for another task.
Multi-task Learning  has  been successful in many scientific application, such as HIV Therapy Screening \cite{bickel2008multi}, analysis of genotype and gene expression data \cite{kim2010tree}, discovery of highly important marker genes \cite{xu2011multi}, modelling of disease progression \cite{zhou2011multi}, disease prediction \cite{zhang2012multi}, biological sequence classification \cite{widmer2010leveraging}, and predicting small interfering RNA (siRNA) efficacy \cite{liu2010multi}. 
Multi-task learning is closely related to the field of transfer learning \cite{thrun1998learning}, in which information is transferred from a specific source task to a specific target task. This can be done by forcing the target model to be structurally or otherwise similar to the source model(s). Neural networks are well suited to transfer learning as both the structure and the model parameters of the source models can be used as a good initializations for the target model, yielding a pre-trained model which can then be further fine-tuned using the available training data on the target task \cite{thrun1994learning, baxter1995learning, bengio2012deep, caruana1995learning}. Especially large image datasets, such as ImageNet \cite{krizhevsky2012imagenet}, have been shown to yield pre-trained models that transfer well to other tasks \cite{donahue2014decaf, sharif2014cnn}. However, it has also been shown that this approach doesn't work well when the target task is not very similar \cite{yosinski2014transferable}. As such, it is often difficult to make transfer learning work for many scientific problems.

The success or failure of multi-task learning often crucially depends on the existence of a good task similarity measure. For instance, one could learn a common Bayesian prior over model parameters trained on multiple tasks and use this to measure between-task similarity \cite{xue2007multi, bakker2003task}, or clustering tasks into groups outright  \cite{jacob2009clustered, argyriou2008convex, evgeniou2005learning}. However, it is usually not straightforward to find a similarity measure that works well.

\section{Transformative Learning}
\begin{figure}
 \begin{center}
  \includegraphics[width=0.9\textwidth]{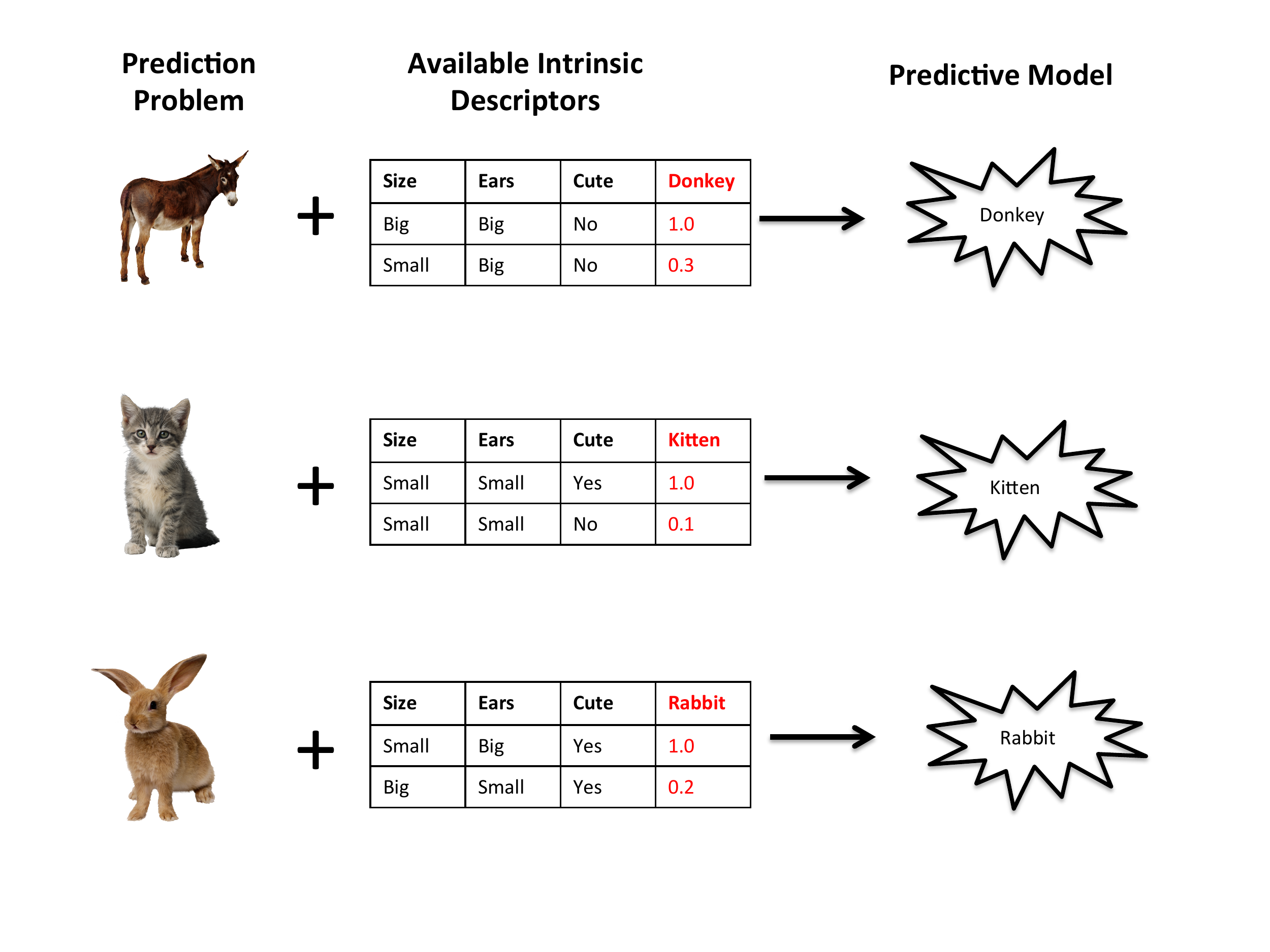}
  \end{center}
\caption{Standard machine learning.  Here there are three related multi-task prediction problems: predicting whether an animal is a donkey or not, kitten or not, or rabbit or not.  Multiple training examples exist for each problem, each described by the same set of intrinsic attributes.  The donkey training examples are used to learn a predictive model for donkeys, the kitten examples to learn a predictive model for kittens, and the rabbit examples to learn a predictive model for rabbits.}
\label{fig:Standard-machine-learning}       % Give a unique label
\end{figure}

\begin{figure}
 \begin{center}
  \includegraphics[width=0.9\textwidth]{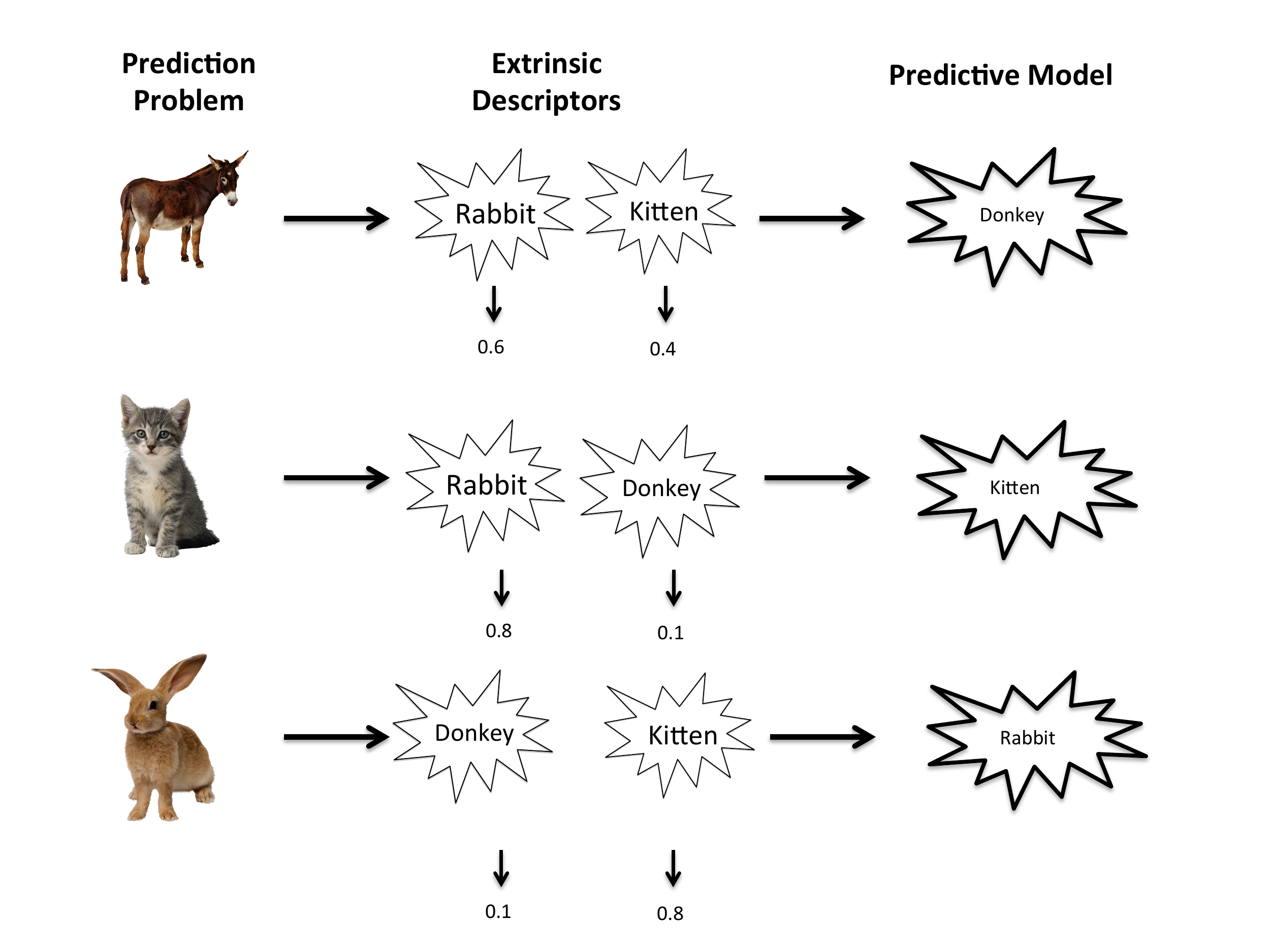}
  \end{center}
\caption{Transformative machine learning.  First three predictive models are learnt for donkeys, kittens, and rabbits, as in standard ML (Fig \ref{fig:Standard-machine-learning}).  Then the kitten and rabbit models are used to predict the donkey examples. For each example in the donkey problem, the kitten model outputs a number, as does the rabbit model.  These numbers are then collected into a tuple, and used as an extrinsic description of the example: the transformed description.  ML is then used to learn a model that classifies examples as donkeys or not.  This process is repeated for the kitten and rabbit problems.}
\label{fig:Transformative-machine-learning}       % Give a unique label
\end{figure}

We present transformative learning a novel method for transforming input representations into more effective ones. The fundamental new idea is to convert a representation based on intrinsic properties to an extrinsic representation based on the predictions on a set of pre-trained models, each trained on another tasks. This leverages available data from many related tasks to perform a combination of multi-task and transfer learning able to make predictions. Transformative learning has the dual advantages of enabling better predictions, and providing explainable explanations.
The input to transformative learning is: (1) a set of related prediction problems, and (2) a set of related examples that have been applied to one or more of the prediction problems. Transformative learning is performed in two learning stages.  In the first learning stage (Fig. \ref{fig:Standard-machine-learning}), separate prediction models are learned for each problem, using the available examples, and their standard intrinsic attributes to describe the examples, producing $n$ predictive models. In the second learning stage (Fig. \ref{fig:Transformative-machine-learning}), for each problem, the available examples are applied to the $n-1$ models to produce $n-1$ predictive values. These values form the transformed representation. Instead of representing examples by intrinsic attributes, they are represented by what other models predict about them.  This transformed extrinsic representation is used to learn the final predictive model.  
In transformative learning we learn task similarity and a joint representation at the same time. Instead of using a predefined similarity measure to pre-select a set of similar tasks, we project the different tasks into one joint numeric representation, and use a meta-learning algorithm to learn from this new representation how to make accurate predictions for the task at hand.

To demonstrate the utility of transformative learning we have applied it to three real-world scientific problems: drug-design (quantitative structure activity relationship learning), predicting human gene expression (across different tissue types and drug treatments), and meta-machine learning (predicting how well machine learning method will work on problems).

\section{Quantitative Structure Activity Relationship Learning}
\label{sec:qsar_learning}
The standard Quantitative Structure Activity Relationship (QSAR) learning problem is: given a target (usually a protein) and a set of chemical compounds (small molecules) with associated bioactivities (normally inhibiting a target protein), learn a predictive mapping from molecular representation to activity.  QSAR problems are suitable for transformative learning as they can be related by having related targets proteins (e.g. the problem of inhibiting mouse DHFR is similar to that of inhibiting human DHFR), and they can also be related by involving the same or chemically related small molecules.

\subsection{Background}
\label{subsec:qsar_background}
Drug development is one of the most important applications of science. It is an essential step in the treatment of almost all diseases. Developing a new drug is however slow and expensive. The average cost to bring a new drug to market is $>2.5$ billion US dollars \cite{mullard2014new}. A key step in drug development is learning QSARs \cite{martin2010tautomerism, cherkasov2014qsar, cumming2013chemical}. Almost every form of statistical and machine learning method has been applied to this problem, but no single method has been found to be always best \cite{olier2018meta}. The most important QSAR dataset is the ChEMBL database \cite{gaulton2016chembl}, a medicinal chemistry database managed by the European Bioinformatics Institute (EBI). It is abstracted and curated from the scientific literature, and covers a significant fraction of the medicinal chemistry corpus. The data consist of information on the drug targets, the structures of the tested compounds (from which different intrinsic chemoinformatic representations may be calculated), and the bioactivities of the compounds on their targets. We extracted 2,219 targets from ChEMBL with a diverse number of chemical compounds, ranging from 30 to about 6,000, each target resulting in a dataset with as many examples as compounds \cite{olier2018meta}. Chemical compounds were intrinsically described using a standard fingerprint representation (as it is the most commonly used in QSAR learning), where the presence or absence of a particular molecular substructure in a molecule (e.g. methyl group, benzene ring) is indicated by a Boolean variable. Specifically, we calculated the 1024 bits FCFP4 fingerprint representation using the Pipeline Pilot software from BIOVIA \cite{rogers2010extended}.

\subsection{Results}
\label{subsec:qsar_results}
We applied transformative learning to generate extrinsic descriptors of the chemical compounds. For this we selected two learning methods: Random Forest (RF, 500 trees) \cite{breiman2001random}, and Linear Regression with Ridge Penalization (Ridge, L2 = 10) \cite{hoerl1970ridge}. This choice was based on the results from \cite{olier2018meta}, where these two methods performed best for QSAR datasets using the 1,024 fingerprint representation. QSAR models were created, one for each dataset and learner. Then extrinsic descriptors were generated by predicting activity using all the models but excluding the one from compound was part of the training set. Therefore, 2,218 extrinsic descriptors were generated per chemical compound (i.e. 2,219 original datasets - 1 training dataset). 
We performed a comparative assessment of the two QSAR data representations: the original intrinsic one based of molecular fingerprints, and the transformed data representation based on model predictions. For the comparison we applied three machine learning methods: Random Forest (RF, 500 trees), Linear Regression with Ridge Penalization (Ridge, L2 = 10), and  Support Vector Machines (SVM, radial basis function kernel, width = 0.2) \cite{cortes1995support}. Method performance was measured using the root mean squared error (RMSE). RMSE, which values are in the same range as the response variable, is standard for regression tasks. 10-fold cross-validation was used across all experiments, with the same data splits to reduce bias risk.  All the experiments were performed in R \cite{team2013r}. Table \ref{tab:QSAR-results} reports average RMSE performance on the test sets. 

\begin{table}
\caption{QSAR Transformative Learning Results. Performance results as measured using the average RMSE after 10-fold cross-validation. In the table: 'Original rep.', 'TL - RF', and 'TL - Ridge' indicate performance using the original intrinsic data representation, transformed representation using random forest, and using ridge penalization, respectively; (\%) indicates the performance improvement of each transformed representation, and is measured as $RMSE_{original} - RMSE_{TL} / RMSE_{original} * 100\% $}
\label{tab:QSAR-results}      
\begin{tabular}{*{6}{p{1.55cm}}}
\hline\noalign{\smallskip}
Learning Method & Original rep. & TL - RF & (\%) & TL - Ridge & (\%) \\
\noalign{\smallskip}\hline\noalign{\smallskip}
RF & 0.1643 & 0.1478 & 10.05 & 0.1642 & 0.06 \\
Ridge & 0.1654 & 0.1655 & -0.06 & 0.1701 & -2.84 \\
SVM & 0.1693 & 0.1522 & 10.10 & 0.1693 & 0.00 \\
\noalign{\smallskip}\hline
\end{tabular}
\end{table}

First considering the application of Random Forest learning to transform the intrinsic chemical representation.  Applying Random Forest learning a second time to the transformed representation was found to outperform the first  Random Forest on 1,118 of the 2,212 problems. This corresponds to $>10$\% mean improvement in RMSE.  A similar result was found applying SVM to this transformed representation where SVM outperform the first  SVM on 1,125 of the 2,212 problems, which also corresponds to a $>$10\% mean improvement in RMSE.  These results are especially noteworthy as we know from previous work, where we compared 18 common learning methods with 3 different intrinsic representations on the same data, that Random Forest with the fingerprint representation is the best method / intrinsic representation combination \cite{olier2018meta}.  Therefore, transformative learning has produced a large improvement over the best of 54 (18 x 6) intrinsic approaches.  
The transformed learning approach does not work well with Linear Regression with Ridge Penalization.  Using Ridge Penalization as the learning method to transform the representation produces no improvement.  Nor is Ridge Penalization successful at exploiting the transformed representation generated by random Forest.  

\section{Gene Expression Learning}
\label{sec:gene_learning}
As our second problem domain we selected the problem of predicting gene expression level. Our goal was to build a predictive models that given a drug and cancer cell type would be able to predict gene expression levels. These models can then be used to guide laboratory-based drug discovery experiments. Specifically, we utilized the Library of Integrated Network-based Cellular Signatures data (LINCS) \cite{koleti2017data}.  This data describes the effect of drugs in cancer cell lines on the expression levels of 978 landmark human genes. 
The prediction problem is to learn models for each gene (978 models) that predict the gene’s expression level, given experimental conditions (cell type, drug, dosage), the related examples are the experimental condition (cell type, drug, dosage).

\subsection{Background}
\label{subsec:gene_background}
We used LINCS Phase II data (accession code GSE70138), which consists of 118,050 experimental conditions, along with the corresponding expression levels for 978 landmark genes. We generated attributes for each perturbation condition using the accompanying metadata. Each experimental condition is associated with a perturbagen (drug), cell type and site, perturbagen dosage, and perturbagen time frame. In total, there are 30 cell types (ct), 14 cell sites (cs), 83 dosages (d) and 3 time points (tp). Of the 2,170 drugs in the dataset, 1,795 have valid chemical structures (canonical smiles codes) according to the metadata. We converted the canonical smiles to the a 1,024 bit FCFP4 fingerprints (fp) using RDKit \cite{landrum2016rdkit}. For all perturbation conditions with valid canonical smiles as rows, we generated Boolean features with the following columns: $[ct_1 \dots ct_{30}][cs_1 \dots cs_{14}][d_1 \dots d_{83}][tp_1 \dots tp_3][fp_1 \dots fp_{1024}]$. This generated a 107,152 × 1,155 experimental condition matrix, row and column identifiers included, which can be used as input for building models to predict the expression levels of the 978 genes using traditional machine learning techniques. For each gene we generated both a train and test set with 7000 and 3000 samples respectively. We did this by first randomly splitting the original perturbation condition data with 107,152 samples and their corresponding gene expression levels, into train and test sets of 70\% and 30\% respectively. Using this main train and test set, we randomly sampled train and test individuals for each gene. The gene expression levels for the 978 genes were normalised such that their values lie between 0.0 and 1.0. We used two learning algorithms for these experiments, Random Forests (RF) and Linear Regression with Ridge Penalization  (Ridge). For the RF 500 trees were grown, a third of the total number of variables were considered at each split, and five observations were used in each terminal node. For Ridge the regularization parameter was chosen using 10-fold internal cross-validation. All the experiments were performed in R \cite{team2013r}. The RF experiments were performed using version 4.6-12 of the randomForest package, and the Ridge experiments were performed using version 2.0-13 of the GLMNET package. Model performance was calculated as the RMSE.
For both, Random Forests and Ridge, we considered 500 descriptors in the transformative learning step. For both learning methods the same gene models were used in the generation of the first order descriptors. 

\subsection{Results}
\label{subsec:gene_results}
First considering the application of Random Forest learning to learn from the intrinsic representation.  Applying Random Forest learning a second time to this transformed representation was found to outperform the first Random Forest on 977 of the 978 genes. This corresponds to a $>4$\% mean improvement in RMSE.  In contrast, applying Ridge learning to the transformed representation was found to outperform the first Random Forest on 862 of the 978 genes. This corresponds to a $>2$\% mean improvement in RMSE, see Table \ref{tab:gene-results}.

\begin{table}
\caption{Gene Expression Transformative Learning Results. Performance results were measured using the RMSE on a test set. Column names as in Table \ref{tab:QSAR-results}.}
\label{tab:gene-results}       
\begin{tabular}{*{6}{p{1.55cm}}}
\hline\noalign{\smallskip}
Learning Method & Original rep. & TL - RF & (\%) & TL - Ridge & (\%) \\
\noalign{\smallskip}\hline\noalign{\smallskip}
RF & 0.0694 & 0.0664 & 4.32 & 0.0675 & 2.74 \\
Ridge & 0.0724 & 0.0673 & 7.04 & 0.0726 & -0.27 \\
\noalign{\smallskip}\hline
\end{tabular}
\end{table}

Then considering the application of Ridge learning to learn from the intrinsic representation. Applying Random Forest learning to this transformed representation was found to outperform the base Ridge models on 952 of the 978 genes.  This corresponds to a $>7$\% mean improvement in RMSE, see Table \ref{tab:gene-results}. In contrast applying Ridge learning to the Ridge learning transformed representation outperformed Ridge learning on only 415 of the 978 genes.

\section{Meta-Learning for Machine Learning}
\label{sec:meta-lrn}
In machine learning, a key challenge is to select the best algorithm to train a predictive model on a new task. One approach to this problem is to apply machine learning itself to predict the best techniques \cite{vanschoren2018meta}.  Hence, this is called meta-learning, and we select it as our third problem domain. In this type of meta-learning, the prediction problem is to predict the performance of a machine learning method (given an exact configuration) on a new task, given the characteristics of the training data (e.g. statistics of the training data distribution). Domain problems can be related by having similar data distributions, data defects (e.g. missing values), or by containing data being generated by similar processes. The properties used to describe the datasets themselves are typically called meta-features.

\subsection{Background}
\label{subsec:metalrn_background}
Meta-learning for machine learning is feasible thanks to the creation of open repositories that collect datasets, meta-features, and experiment results. OpenML is an online machine learning platform where researchers can automatically log and share data, code, and experiments \cite{vanschoren2014openml}. It brings together reproducible experiments from most major machine learning environments, such as WEKA (Java), mlr (R), and scikit-learn (Python). From OpenML we retrieved data from an earlier meta-learning study.\footnote{Details can be found on \url{https://www.openml.org/s/7}.} Although we had to exclude a few tasks and algorithms because they lacked sufficient evaluations in OpenML, this yielded a set of 10840 evaluations on 351 tasks (datasets) and 53 machine learning methods (called flows on OpenML) from mlr \cite{mlr2017}. From each task, 21 dataset descriptors were extracted, such as the number of examples, number of missing values, and percentage of numeric features. We formed meta-datasets, one for each machine learning method. An observation within a meta-dataset represents an original OpenML task, and each feature, a dataset descriptor. The original aim of the study was to predict the area under the ROC (AUC). Therefore, in total, we produced 53 meta-datasets with a diverse number of OpenML tasks, ranging from above 100 to about 250.
We applied transformative learning to transform the original representation of the datasets into extrinsic descriptors of the OpenML tasks. Three learners were selected to do the transformation: Random Forest (RF, 500 trees), Linear Regression with Ridge Penalization (Ridge, L2 = 10), and Support Vector Machines with Radial Basis Kernel Functions (SVM, $\sigma = 0.2$). The transformed descriptors were generated by predicting AUC using all available models but excluding the one from the which the OpenML task belonged. In this way 52 extrinsic descriptors were generated for each OpenML task.

\subsection{Results}
Table \ref{tab:metalrn-results} shows comparative performance results between the two data representation: the intrinsic original representation using data descriptors (i.e. number of instances, percentage of numeric features, etc), and the transformed extrinsic representation. We used similar learners as above (RF, 500 trees; Ridge, L=10; and SVM, $\sigma=0.2$). For instance, when we train a Random Forest on the intrinsic representation and use it to predict the performance of learning algorithms on every dataset, those predictions have an RMSE of 0.1184 (first row in Table \ref{tab:metalrn-results}).  Training the Random Forest learning on the transformed representation (which does not have access to the dataset we are predicting for) was found to outperform the first Random Forest on 51 of the 52 tasks, and yielding an RMSE of 0.526. This corresponds to an impressive $>55\%$ mean improvement in RMSE.  Similarly, applying Ridge to the transformed representation was found to outperform the first Random Forest on all of the 52 tasks, which corresponds to $>49\%$ mean improvement in RMSE.  Applying SVM to the transformed representation was found to outperform the first Random Forest on 50 of the 52 tasks, which corresponds to $>57\%$ mean improvement in RMSE.  
Likewise, applying an SVM to learn from the transformed representations was found to vastly outperform training on the intrinsic representation (third row in Table \ref{tab:metalrn-results}),  corresponding to $>27\%$ mean improvement in RMSE. Learning on features transformed by the Random Forest learning was found to outperform the original SVM model on 50 of the 52 tasks and $>20\%$ mean improvement in RMSE, and using features transformed by Ridge was found to outperform the first SVM method on all of the 52 tasks, which corresponds to $>25\%$ mean improvement in RMSE.  

\begin{table}
\caption{Meta-learning for Machine Learning Transformative Learning Results. Performance results as measured using the average RMSE after 10-fold cross-validation. Column names follow same naming as in Table \ref{tab:QSAR-results}.}
\label{tab:metalrn-results}      
\begin{tabular}{*{3}{p{1.25cm}}lp{1.4cm}lp{1.3cm}l}
\hline\noalign{\smallskip}
Learning Method & Original rep. & TL - RF & (\%) & TL - Ridge & (\%) & TL - SVM & (\%) \\
\noalign{\smallskip}\hline\noalign{\smallskip}
RF & 0.1184 & 0.0526 & 55.57 & 0.1236 & -4.39 & 0.0939 & 20.69 \\
Ridge & 0.1403 & 0.0710 & 49.39 & 0.1356 & 3.35 & 0.1047 & 25.37 \\
SVM & 0.1335 & 0.0573 & 57.08 & 0.1352 & -1.27 & 0.0972 & 27.19 \\
\noalign{\smallskip}\hline
\end{tabular}
\end{table}

As with QSAR learning and gene expression prediction the application of Ridge learning to transform the representation was unsuccessful, with results little different from the original intrinsic representation. 

\section{Discussion}
\label{sec:discussion}
\paragraph{Comparison with All-In Learning.}
A standard meta-learning approach, often used with DNNs, is to try to learn one large model that encompasses all the problems.  In some circumstances this can work well.  However, this approach has clear disadvantages compared to transformative learning:
\begin{itemize}
    \item If new data occurs for a task, the whole model has to be relearned.
    \item If a new task is added, the whole model has to be relearned.
    \item The relationships between tasks are not explicit.
    \item The relationships between examples are also not explicit.

\end{itemize}
 
\paragraph{Explainable AI.}
A major motivation of transformative learning is to develop a learning approach that provides explainable models.  The transformed representation generates clearly understandable descriptors for learning.  For example, using the example problem in Fig \ref{fig:Transformative-machine-learning} of classifying animals, it is possible to classify an animal as a rabbit if it has a combination of properties of a donkey and kitten. This explainability is in marked contrast to the black-box nature of DNNs. 
Transformative learning also enables one to better understand the relationships between the learning tasks.  This can be achieved by using the models for each task to predict all the examples, and then clustering the tasks by their predictions: which displays how the tasks are related in prediction space.  Similarly, it is possible to better understand the relationships between examples by clustering them by their different model predictions:  which shows how the examples are related in task space. 
 
\paragraph{The Computational Cost of Transformative Learning.}
One disadvantage of transformative learning is its additional computational cost. With transformative learning, in addition to the standard learning process, it is necessary to: 1) use each task model to predict all the examples to form the transformed representation, and 2) learn new task models using the transformed representation.  Both tasks are potentially computationally expensive.  However, the cost of transformative learning is low compared to DNNs.
 
\paragraph{Transformative Learning using Linear Regression with Ridge Penalization.}
Our results indicate that the use of Ridge to form a transformed representation does not result in improved predictions.  This suggests that it is necessary for the learning method that forms the transformed representation to be non-linear.  In contrast, the use of Ridge to make predictions based on the a transformed representation made by Random Forests and SVM can work well, as it does for Gene Expression prediction and Meta-learning for Machine Learning.
 
\paragraph{Second-Order Transformative Learning.}
In transformative learning the fundamental new idea is to transform the original, intrinsic data representation, to an extrinsic representation based on what a pre-trained set of models predict about the examples. Given the expectation that using the transformed representation produces better predictions than the original intrinsic representation, it is natural to extend the idea of transformative learning by applying it a second time, i.e. to use the predictions from the transformed representation to form a second-order transformed representation.  As the predictions from the transformed representation are better than the ones from intrinsic representation, learning using second-order transformed representation should be more successful than with the first -order transformed representation.  One clear disadvantage with this approach is the high-computational cost of using a second-order transformed representation. 
 
\section{Conclusions}
\label{conclusions}
In the past, machine learning was most commonly applied in bespoke ways to isolated problems.  Now, with the ever-increasing availability of data, machine learning is being increasingly applied to large sets of related problems.  This is motivating an increased interest in multi-task and transfer learning.  We have developed a novel and general representation learning approach for multi-task learning, and we have demonstrated the success of this approach on three real-world scientific problems: drug-design, predicting human gene expression, and meta-learning for machine learning. In all three problems, transformative machine learning significantly outperforms the best intrinsic representations. We expect transformative learning to be of general application to scientific problems and beyond.

\begin{acknowledgements}
The authors would like to thank Rafael Mantovani for generating the original meta-learning data used in this study.

\end{acknowledgements}

\bibliographystyle{abbrvnat}      % basic style, author-year citations
\bibliography{references}   % name your BibTeX data base

\begin{thebibliography}{49}
\providecommand{\natexlab}[1]{#1}
\providecommand{\url}[1]{\texttt{#1}}
\expandafter\ifx\csname urlstyle\endcsname\relax
  \providecommand{\doi}[1]{doi: #1}\else
  \providecommand{\doi}{doi: \begingroup \urlstyle{rm}\Url}\fi

\bibitem[Ando and Zhang(2005)]{ando2005framework}
R.~K. Ando and T.~Zhang.
\newblock A framework for learning predictive structures from multiple tasks
  and unlabeled data.
\newblock \emph{Journal of Machine Learning Research}, 6\penalty0
  (Nov):\penalty0 1817--1853, 2005.

\bibitem[Argyriou et~al.(2008)Argyriou, Evgeniou, and
  Pontil]{argyriou2008convex}
A.~Argyriou, T.~Evgeniou, and M.~Pontil.
\newblock Convex multi-task feature learning.
\newblock \emph{Machine Learning}, 73\penalty0 (3):\penalty0 243--272, 2008.

\bibitem[Bakker and Heskes(2003)]{bakker2003task}
B.~Bakker and T.~Heskes.
\newblock Task clustering and gating for bayesian multitask learning.
\newblock \emph{Journal of Machine Learning Research}, 4\penalty0
  (May):\penalty0 83--99, 2003.

\bibitem[Baxter(1995)]{baxter1995learning}
J.~Baxter.
\newblock Learning internal representations.
\newblock In \emph{Proceedings of the eighth annual conference on Computational
  learning theory}, pages 311--320. ACM, 1995.

\bibitem[Bengio(2012)]{bengio2012deep}
Y.~Bengio.
\newblock Deep learning of representations for unsupervised and transfer
  learning.
\newblock In \emph{Proceedings of ICML Workshop on Unsupervised and Transfer
  Learning}, pages 17--36, 2012.

\bibitem[Bickel et~al.(2008)Bickel, Bogojeska, Lengauer, and
  Scheffer]{bickel2008multi}
S.~Bickel, J.~Bogojeska, T.~Lengauer, and T.~Scheffer.
\newblock Multi-task learning for hiv therapy screening.
\newblock In \emph{Proceedings of the 25th international conference on Machine
  learning}, pages 56--63. ACM, 2008.

\bibitem[Bischl et~al.(2016)Bischl, Lang, Kotthoff, Schiffner, Richter,
  Studerus, Casalicchio, and Jones]{mlr2017}
B.~Bischl, M.~Lang, L.~Kotthoff, J.~Schiffner, J.~Richter, E.~Studerus,
  G.~Casalicchio, and Z.~M. Jones.
\newblock mlr: Machine learning in r.
\newblock \emph{Journal of Machine Learning Research}, 17\penalty0
  (170):\penalty0 1--5, 2016.
\newblock URL \url{http://jmlr.org/papers/v17/15-066.html}.

\bibitem[Breiman(2001)]{breiman2001random}
L.~Breiman.
\newblock Random forests.
\newblock \emph{Machine learning}, 45\penalty0 (1):\penalty0 5--32, 2001.

\bibitem[Buchanan et~al.(1968)Buchanan, Sutherland, and
  Feigenbaum]{buchanan1968heuristic}
B.~Buchanan, G.~Sutherland, and E.~A. Feigenbaum.
\newblock \emph{Heuristic DENDRAL: A program for generating explanatory
  hypotheses in organic chemistry}.
\newblock Stanford University, 1968.

\bibitem[Butler et~al.(2018)Butler, Davies, Cartwright, Isayev, and
  Walsh]{butler2018machine}
K.~T. Butler, D.~W. Davies, H.~Cartwright, O.~Isayev, and A.~Walsh.
\newblock Machine learning for molecular and materials science.
\newblock \emph{Nature}, 559\penalty0 (7715):\penalty0 547, 2018.

\bibitem[Caruana(1995)]{caruana1995learning}
R.~Caruana.
\newblock Learning many related tasks at the same time with backpropagation.
\newblock In \emph{Advances in neural information processing systems}, pages
  657--664, 1995.

\bibitem[Caruana(1997)]{caruana1997multitask}
R.~Caruana.
\newblock Multitask learning.
\newblock \emph{Machine learning}, 28\penalty0 (1):\penalty0 41--75, 1997.

\bibitem[Cherkasov et~al.(2014)Cherkasov, Muratov, Fourches, Varnek, Baskin,
  Cronin, Dearden, Gramatica, Martin, Todeschini, et~al.]{cherkasov2014qsar}
A.~Cherkasov, E.~N. Muratov, D.~Fourches, A.~Varnek, I.~I. Baskin, M.~Cronin,
  J.~Dearden, P.~Gramatica, Y.~C. Martin, R.~Todeschini, et~al.
\newblock Qsar modeling: where have you been? where are you going to?
\newblock \emph{Journal of medicinal chemistry}, 57\penalty0 (12):\penalty0
  4977--5010, 2014.

\bibitem[Cortes and Vapnik(1995)]{cortes1995support}
C.~Cortes and V.~Vapnik.
\newblock Support-vector networks.
\newblock \emph{Machine learning}, 20\penalty0 (3):\penalty0 273--297, 1995.

\bibitem[Cumming et~al.(2013)Cumming, Davis, Muresan, Haeberlein, and
  Chen]{cumming2013chemical}
J.~G. Cumming, A.~M. Davis, S.~Muresan, M.~Haeberlein, and H.~Chen.
\newblock Chemical predictive modelling to improve compound quality.
\newblock \emph{Nature reviews Drug discovery}, 12\penalty0 (12):\penalty0 948,
  2013.

\bibitem[Donahue et~al.(2014)Donahue, Jia, Vinyals, Hoffman, Zhang, Tzeng, and
  Darrell]{donahue2014decaf}
J.~Donahue, Y.~Jia, O.~Vinyals, J.~Hoffman, N.~Zhang, E.~Tzeng, and T.~Darrell.
\newblock Decaf: A deep convolutional activation feature for generic visual
  recognition.
\newblock In \emph{International conference on machine learning}, pages
  647--655, 2014.

\bibitem[Esteva et~al.(2017)Esteva, Kuprel, Novoa, Ko, Swetter, Blau, and
  Thrun]{esteva2017dermatologist}
A.~Esteva, B.~Kuprel, R.~A. Novoa, J.~Ko, S.~M. Swetter, H.~M. Blau, and
  S.~Thrun.
\newblock Dermatologist-level classification of skin cancer with deep neural
  networks.
\newblock \emph{Nature}, 542\penalty0 (7639):\penalty0 115, 2017.

\bibitem[Evgeniou et~al.(2005)Evgeniou, Micchelli, and
  Pontil]{evgeniou2005learning}
T.~Evgeniou, C.~A. Micchelli, and M.~Pontil.
\newblock Learning multiple tasks with kernel methods.
\newblock \emph{Journal of Machine Learning Research}, 6\penalty0
  (Apr):\penalty0 615--637, 2005.

\bibitem[Gaulton et~al.(2016)Gaulton, Hersey, Nowotka, Bento, Chambers, Mendez,
  Mutowo, Atkinson, Bellis, Cibri{\'a}n-Uhalte, et~al.]{gaulton2016chembl}
A.~Gaulton, A.~Hersey, M.~Nowotka, A.~P. Bento, J.~Chambers, D.~Mendez,
  P.~Mutowo, F.~Atkinson, L.~J. Bellis, E.~Cibri{\'a}n-Uhalte, et~al.
\newblock The chembl database in 2017.
\newblock \emph{Nucleic acids research}, 45\penalty0 (D1):\penalty0 D945--D954,
  2016.

\bibitem[Hoerl and Kennard(1970)]{hoerl1970ridge}
A.~E. Hoerl and R.~W. Kennard.
\newblock Ridge regression: Biased estimation for nonorthogonal problems.
\newblock \emph{Technometrics}, 12\penalty0 (1):\penalty0 55--67, 1970.

\bibitem[Jacob et~al.(2009)Jacob, Vert, and Bach]{jacob2009clustered}
L.~Jacob, J.-p. Vert, and F.~R. Bach.
\newblock Clustered multi-task learning: A convex formulation.
\newblock In \emph{Advances in neural information processing systems}, pages
  745--752, 2009.

\bibitem[Kim and Xing(2010)]{kim2010tree}
S.~Kim and E.~P. Xing.
\newblock Tree-guided group lasso for multi-task regression with structured
  sparsity.
\newblock In \emph{ICML}, pages 543--550, 2010.

\bibitem[King et~al.(2009)King, Rowland, Oliver, Young, Aubrey, Byrne, Liakata,
  Markham, Pir, Soldatova, et~al.]{king2009automation}
R.~D. King, J.~Rowland, S.~G. Oliver, M.~Young, W.~Aubrey, E.~Byrne,
  M.~Liakata, M.~Markham, P.~Pir, L.~N. Soldatova, et~al.
\newblock The automation of science.
\newblock \emph{Science}, 324\penalty0 (5923):\penalty0 85--89, 2009.

\bibitem[Koleti et~al.(2017)Koleti, Terryn, Stathias, Chung, Cooper, Turner,
  Vidovi{\'c}, Forlin, Kelley, D’Urso, et~al.]{koleti2017data}
A.~Koleti, R.~Terryn, V.~Stathias, C.~Chung, D.~J. Cooper, J.~P. Turner,
  D.~Vidovi{\'c}, M.~Forlin, T.~T. Kelley, A.~D’Urso, et~al.
\newblock Data portal for the library of integrated network-based cellular
  signatures (lincs) program: integrated access to diverse large-scale cellular
  perturbation response data.
\newblock \emph{Nucleic acids research}, 46\penalty0 (D1):\penalty0 D558--D566,
  2017.

\bibitem[Krizhevsky et~al.(2012)Krizhevsky, Sutskever, and
  Hinton]{krizhevsky2012imagenet}
A.~Krizhevsky, I.~Sutskever, and G.~E. Hinton.
\newblock Imagenet classification with deep convolutional neural networks.
\newblock In \emph{Advances in neural information processing systems}, pages
  1097--1105, 2012.

\bibitem[Landrum(2016)]{landrum2016rdkit}
G.~Landrum.
\newblock Rdkit: open-source cheminformatics http://www. rdkit. org, 2016.

\bibitem[LeCun et~al.(2015)LeCun, Bengio, and Hinton]{lecun2015deep}
Y.~LeCun, Y.~Bengio, and G.~Hinton.
\newblock Deep learning.
\newblock \emph{nature}, 521\penalty0 (7553):\penalty0 436, 2015.

\bibitem[Liu et~al.(2010)Liu, Xu, Zheng, Xue, Cao, and Yang]{liu2010multi}
Q.~Liu, Q.~Xu, V.~W. Zheng, H.~Xue, Z.~Cao, and Q.~Yang.
\newblock Multi-task learning for cross-platform sirna efficacy prediction: an
  in-silico study.
\newblock \emph{BMC bioinformatics}, 11\penalty0 (1):\penalty0 181, 2010.

\bibitem[Martin(2010)]{martin2010tautomerism}
Y.~C. Martin.
\newblock Tautomerism, hammett $\sigma$, and qsar.
\newblock \emph{Journal of computer-aided molecular design}, 24\penalty0
  (6-7):\penalty0 613--616, 2010.

\bibitem[Mullard(2014)]{mullard2014new}
A.~Mullard.
\newblock New drugs cost us \$2.6 billion to develop, 2014.

\bibitem[Olier et~al.(2018)Olier, Sadawi, Bickerton, Vanschoren, Grosan,
  Soldatova, and King]{olier2018meta}
I.~Olier, N.~Sadawi, G.~R. Bickerton, J.~Vanschoren, C.~Grosan, L.~Soldatova,
  and R.~D. King.
\newblock Meta-qsar: a large-scale application of meta-learning to drug design
  and discovery.
\newblock \emph{Machine Learning}, 107\penalty0 (1):\penalty0 285--311, 2018.

\bibitem[Rogers and Hahn(2010)]{rogers2010extended}
D.~Rogers and M.~Hahn.
\newblock Extended-connectivity fingerprints.
\newblock \emph{Journal of chemical information and modeling}, 50\penalty0
  (5):\penalty0 742--754, 2010.

\bibitem[Russell and Norvig(2016)]{russell2016artificial}
S.~J. Russell and P.~Norvig.
\newblock \emph{Artificial intelligence: a modern approach}.
\newblock Malaysia; Pearson Education Limited,, 2016.

\bibitem[Schmidt and Lipson(2009)]{schmidt2009distilling}
M.~Schmidt and H.~Lipson.
\newblock Distilling free-form natural laws from experimental data.
\newblock \emph{science}, 324\penalty0 (5923):\penalty0 81--85, 2009.

\bibitem[Schneider(2017)]{schneider2017automating}
G.~Schneider.
\newblock Automating drug discovery.
\newblock \emph{Nature Reviews Drug Discovery}, 17\penalty0 (2):\penalty0 97,
  2017.

\bibitem[Segler et~al.(2018)Segler, Preuss, and Waller]{segler2018planning}
M.~H. Segler, M.~Preuss, and M.~P. Waller.
\newblock Planning chemical syntheses with deep neural networks and symbolic
  ai.
\newblock \emph{Nature}, 555\penalty0 (7698):\penalty0 604, 2018.

\bibitem[Sharif~Razavian et~al.(2014)Sharif~Razavian, Azizpour, Sullivan, and
  Carlsson]{sharif2014cnn}
A.~Sharif~Razavian, H.~Azizpour, J.~Sullivan, and S.~Carlsson.
\newblock Cnn features off-the-shelf: an astounding baseline for recognition.
\newblock In \emph{Proceedings of the IEEE conference on computer vision and
  pattern recognition workshops}, pages 806--813, 2014.

\bibitem[Silver et~al.(2016)Silver, Huang, Maddison, Guez, Sifre, Van
  Den~Driessche, Schrittwieser, Antonoglou, Panneershelvam, Lanctot,
  et~al.]{silver2016mastering}
D.~Silver, A.~Huang, C.~J. Maddison, A.~Guez, L.~Sifre, G.~Van Den~Driessche,
  J.~Schrittwieser, I.~Antonoglou, V.~Panneershelvam, M.~Lanctot, et~al.
\newblock Mastering the game of go with deep neural networks and tree search.
\newblock \emph{nature}, 529\penalty0 (7587):\penalty0 484, 2016.

\bibitem[Team et~al.(2013)]{team2013r}
R.~C. Team et~al.
\newblock R: A language and environment for statistical computing.
\newblock 2013.

\bibitem[Thrun and Mitchell(1994)]{thrun1994learning}
S.~Thrun and T.~M. Mitchell.
\newblock Learning one more thing.
\newblock Technical report, CARNEGIE-MELLON UNIV PITTSBURGH PA DEPT OF COMPUTER
  SCIENCE, 1994.

\bibitem[Thrun and Pratt(1998)]{thrun1998learning}
S.~Thrun and L.~Pratt.
\newblock Learning to learn: Introduction and overview.
\newblock In \emph{Learning to learn}, pages 3--17. Springer, 1998.

\bibitem[Vanschoren(2018)]{vanschoren2018meta}
J.~Vanschoren.
\newblock Meta-learning: A survey.
\newblock \emph{arXiv preprint arXiv:1810.03548}, 2018.

\bibitem[Vanschoren et~al.(2014)Vanschoren, Van~Rijn, Bischl, and
  Torgo]{vanschoren2014openml}
J.~Vanschoren, J.~N. Van~Rijn, B.~Bischl, and L.~Torgo.
\newblock {OpenML}: networked science in machine learning.
\newblock \emph{ACM SIGKDD Explorations Newsletter}, 15\penalty0 (2):\penalty0
  49--60, 2014.

\bibitem[Widmer et~al.(2010)Widmer, Leiva, Altun, and
  R{\"a}tsch]{widmer2010leveraging}
C.~Widmer, J.~Leiva, Y.~Altun, and G.~R{\"a}tsch.
\newblock Leveraging sequence classification by taxonomy-based multitask
  learning.
\newblock In \emph{Annual International Conference on Research in Computational
  Molecular Biology}, pages 522--534. Springer, 2010.

\bibitem[Xu et~al.(2011)Xu, Xue, and Yang]{xu2011multi}
Q.~Xu, H.~Xue, and Q.~Yang.
\newblock Multi-platform gene-expression mining and marker gene analysis.
\newblock \emph{International journal of data mining and bioinformatics},
  5\penalty0 (5):\penalty0 485--503, 2011.

\bibitem[Xue et~al.(2007)Xue, Liao, Carin, and Krishnapuram]{xue2007multi}
Y.~Xue, X.~Liao, L.~Carin, and B.~Krishnapuram.
\newblock Multi-task learning for classification with dirichlet process priors.
\newblock \emph{Journal of Machine Learning Research}, 8\penalty0
  (Jan):\penalty0 35--63, 2007.

\bibitem[Yosinski et~al.(2014)Yosinski, Clune, Bengio, and
  Lipson]{yosinski2014transferable}
J.~Yosinski, J.~Clune, Y.~Bengio, and H.~Lipson.
\newblock How transferable are features in deep neural networks?
\newblock In \emph{Advances in neural information processing systems}, pages
  3320--3328, 2014.

\bibitem[Zhang et~al.(2012)Zhang, Shen, Initiative, et~al.]{zhang2012multi}
D.~Zhang, D.~Shen, A.~D.~N. Initiative, et~al.
\newblock Multi-modal multi-task learning for joint prediction of multiple
  regression and classification variables in alzheimer's disease.
\newblock \emph{NeuroImage}, 59\penalty0 (2):\penalty0 895--907, 2012.

\bibitem[Zhou et~al.(2011)Zhou, Yuan, Liu, and Ye]{zhou2011multi}
J.~Zhou, L.~Yuan, J.~Liu, and J.~Ye.
\newblock A multi-task learning formulation for predicting disease progression.
\newblock In \emph{Proceedings of the 17th ACM SIGKDD international conference
  on Knowledge discovery and data mining}, pages 814--822. ACM, 2011.

\end{thebibliography}

\end{document}